\documentclass[letterpaper]{article} 
\usepackage{aaai2026}  
\usepackage{times}  
\usepackage{helvet}  
\usepackage{courier}  
\usepackage[hyphens]{url}  
\usepackage{graphicx} 
\usepackage{natbib}  
\usepackage{caption} 
\usepackage{algorithm}
\usepackage{algorithmic}

\usepackage{newfloat}
\usepackage{listings}
\DeclareCaptionStyle{ruled}{labelfont=normalfont,labelsep=colon,strut=off} 
\floatstyle{ruled}
\newfloat{listing}{tb}{lst}{}
\floatname{listing}{Listing}
\usepackage{amssymb}
\usepackage{amsmath}
\usepackage{multirow}
\usepackage{subcaption}
\usepackage{pifont}
\newcommand{\cmark}{\ding{51}}

\usepackage{colortbl}
\definecolor{mygreen}{rgb}{0.2,0.6,0.6}

\lstdefinestyle{mystyle}{
    backgroundcolor=\color{white},
    basicstyle=\ttfamily\scriptsize,
    commentstyle=\color{mygreen},
    keywordstyle=\color{black},
    numberstyle=\color{gray},
    numbers=none,
    stringstyle=\color{red},
    breaklines=true,
    showstringspaces=false,
    frame=none,
    xleftmargin=0.5em,
    xrightmargin=0.5em,
    aboveskip=0pt,
    belowskip=0pt
}

\title{EM-KD: Distilling Efficient Multimodal Large Language Model with Unbalanced Vision Tokens}
\author{
    Ze Feng\textsuperscript{\rm 1, \rm 2}\thanks{This work was jointly done when ze feng was an intern in Baidu VIS.}, Sen Yang\textsuperscript{\rm 2}, Boqiang Duan\textsuperscript{\rm 2}, Wankou Yang\textsuperscript{\rm 1}\thanks{Corresponding author}, Jingdong Wang\textsuperscript{\rm 2}
}
\affiliations{

    \textsuperscript{\rm 1}Southeast University \\
    \textsuperscript{\rm 2}Baidu Inc \\
    shannon.ze.feng@gmail.com, wkyang@seu.edu.cn
}

\usepackage{bibentry}

\begin{document}

\maketitle

\begin{abstract}
Efficient Multimodal Large Language Models (MLLMs) compress vision tokens to reduce resource consumption, but the loss of visual information can degrade comprehension capabilities.
Although some priors introduce Knowledge Distillation to enhance student models, they overlook the fundamental differences in fine-grained vision comprehension caused by unbalanced vision tokens between the efficient student and vanilla teacher.
In this paper, we propose EM-KD, a novel paradigm that enhances the Efficient MLLMs with Knowledge Distillation.
To overcome the challenge of unbalanced vision tokens, we first calculate the Manhattan distance between the vision logits of teacher and student, and then align them in the spatial dimension with the Hungarian matching algorithm.
After alignment, EM-KD introduces two distillation strategies: 1) Vision-Language Affinity Distillation (VLAD) and 2) Vision Semantic Distillation (VSD).
Specifically, VLAD calculates the affinity matrix between text tokens and aligned vision tokens, and minimizes the smooth L1 distance of the student and the teacher affinity matrices.
Considering the semantic richness of vision logits in the final layer, VSD employs the reverse KL divergence to measure the discrete probability distributions of the aligned vision logits over the vocabulary space.
Comprehensive evaluation on diverse benchmarks demonstrates that EM-KD trained model outperforms prior Efficient MLLMs on both accuracy and efficiency with a large margin, validating its effectiveness.
Compared with previous distillation methods, which are equipped with our proposed vision token matching strategy for fair comparison, EM-KD also achieves better performance.
\end{abstract}


\section{Introduction}

\label{sec:intro}
\begin{figure}[!ht]
\centering
\includegraphics[width=0.95\columnwidth]{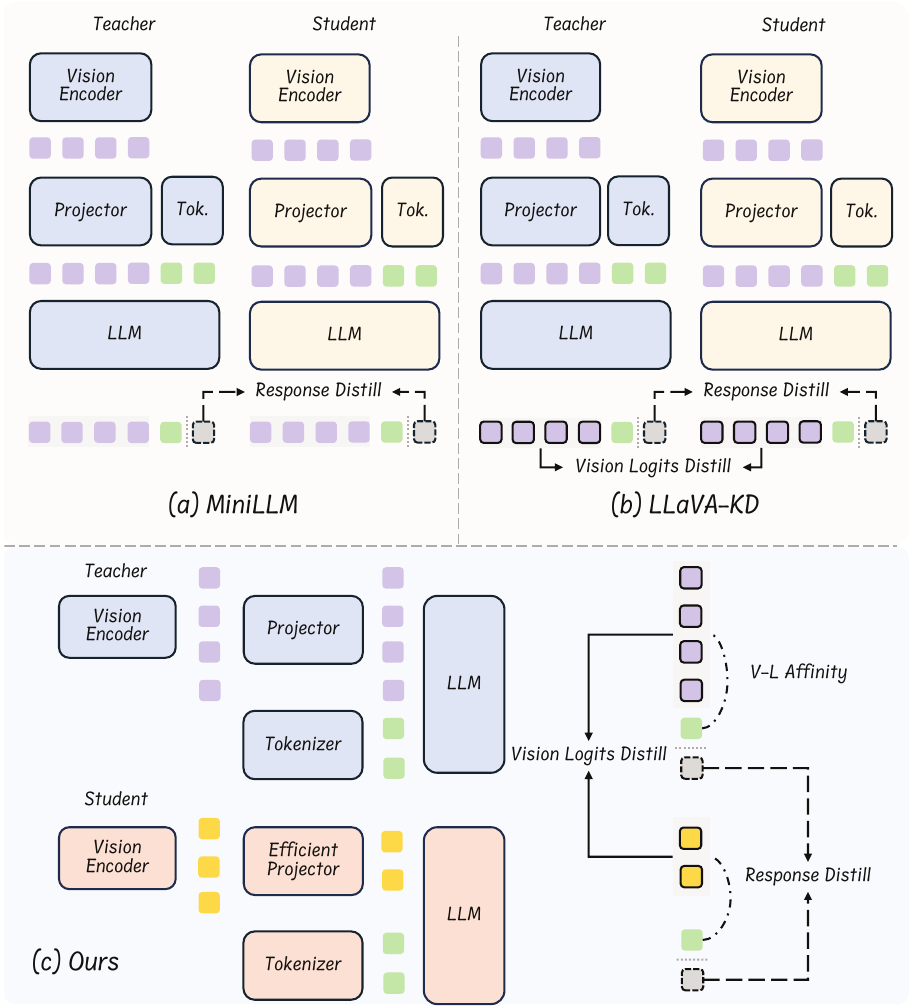} 
\caption{Comparison with existing MLLM distillation approaches. (a) MiniLLM (represent the LLM-style methods) focuses on response token distillation but neglects both the distinctive characteristics of visual features and vision-language correlations. (b) LLaVA-KD (represent the MLLM-style methods) aligns the teacher and
student vision representations by modeling the correlations
between vision tokens, but is limited to the condition of spatial alignment. (c) Our EM-KD can generalize to scenarios with unbalanced vision tokens between teachers and students, and introducing vision-language affinity distillation further enhances cross-modal alignment.}
\label{teaser}
\end{figure}
Multimodal Large Language Models (MLLMs) \cite{bai2023qwenvl, lu2024deepseekvl, liu2024visual, li2023blip, chen2024internvl} have emerged as a cornerstone of artificial intelligence, unifying visual and language understanding to tackle complex tasks such as visual question answering, image captioning, and embodied reasoning. 
While these models demonstrate remarkable versatility, their massive computational demands—stemming from intricate architectures and extensive multimodal pretraining—pose significant barriers to real-world deployment \cite{chen2024image,xing2024pyramiddrop,zhang2025vispruner}, particularly in resource-constrained environments. 
Recent efforts toward Efficient MLLMs \cite{yao2024deco, li2025tokenpacker, chen2024image, feng2025vision} aim to mitigate these challenges by removing the redundant vision tokens or compressing them in vision projectors, leading to higher computational efficiency.
However, reducing vision tokens inevitably leads to information loss, resulting in degraded model performance and generalization, particularly in tasks requiring fine-grained image understanding \cite{feng2025vision}.
Enhancing Efficient MLLMs comprehension without sacrificing inference efficiency remains a challenge that has not been fully explored.

\begin{figure}[t]
\centering
\includegraphics[width=1.00\columnwidth]{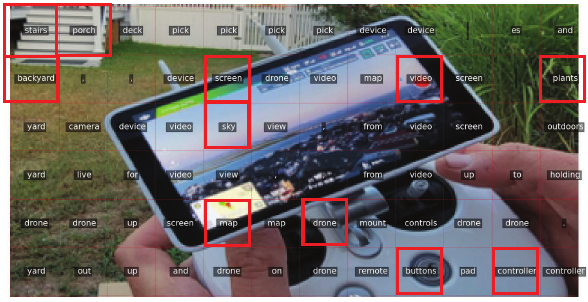} 
\caption{We decode each vision token into vocabulary space via LM head, and find vision logits exhibit rich semantic.}
\label{visualization}
\end{figure}

\begin{figure}[t]
\centering
\includegraphics[width=0.99\columnwidth]{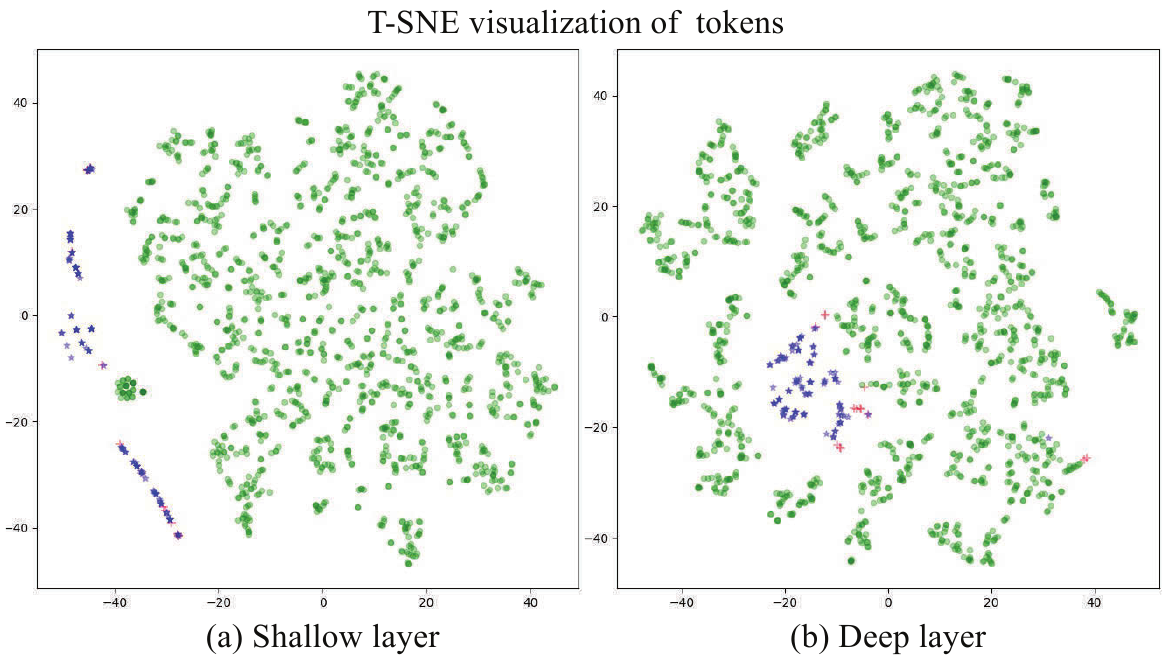} 
\caption{T-SNE visualization of tokens. Vision and textual representations gradually intermix during LLM propagation. Red: system tokens, blue: text tokens, green: vision tokens.}
\label{tsne}
\end{figure}

As a training-phase-only technique, Knowledge Distillation can effectively improve student model capability and performance by incorporating guidance from teacher models or mimicking their behavioral patterns \cite{wang2021knowledge, gou2021knowledge}.
Beyond its prevalent applications in CV and NLP, knowledge distillation has been successfully extended to MLLMs.
LLaVA-KD \cite{cai2025llavakd} proposes a three-stage distillation paradigm which an MLLM is trained with instructions from a larger teacher.
Align-KD \cite{feng2025alignkd} employs an efficient teacher to distill an efficient student model, enhancing the student model’s operational efficiency on edge devices.
The difference between the teacher and student in previous methods lies solely in the size of models, and typically requires some MLPs to align their feature dimensions.
In this work, we employ a well-pretrained vanilla teacher model to distill knowledge into an efficient student model, thereby enhancing the visual understanding and parsing capabilities of Efficient MLLMs. 
However, varying resolutions, vision encoders, and projectors lead to \textit{\textbf{unbalanced vision tokens} between vanilla teacher and efficient student models} (shown in Figure \ref{teaser}).
Previous approaches fail to generalize across such diverse configurations, as they inherently \textit{rely on spatially aligned (\textit{i.e.}, one-to-one correspondences) vision token sequences.}
Since the distillation object plays a critical role in the distillation strategy, here we conduct a preliminary analysis to identify an effective one.
Inspired by \cite{towards}, we employ the language model head (LM head) to decode vision tokens into vocabulary space (\textit{i.e.} vision logits), revealing that image patches often map to semantically meaningful words and final-layer vision logits exhibit \textit{rich semantics} (shown in Figure \ref{visualization}).
Further, t-SNE visualizations of tokens (shown in Figure \ref{tsne}) demonstrate that vision and textual representations gradually intermix during LLM forward.
Our analysis provides two directions: (1) it is feasible to treat vision logits as distillable semantic targets like text logits, and (2) distilling cross-modal relationships could further strengthen alignment between vision and language.

In this paper, we propose EM-KD, a novel paradigm that distills an Efficient MLLM with a well-pretrained teacher.
To address the challenge of unbalanced vision tokens, we consider a set matching task and employ the Hungarian algorithm to establish optimal token-level correspondences between the teacher and student vision tokens.
In order to achieve optimal token-wise distillation, we measure pairwise distances of vision logits to determine the best matches, followed by one-to-one knowledge transfer between aligned teacher-student pairs.
Considering the semantic richness of vision logits in the final layer, EM-KD introduces Vision Semantic Distillation, which employs the reverse KL divergence to measure the distance between discrete probability distributions over the vocabulary space, rather than relying on hidden-level token similarity.
Furthermore, we propose distilling the affinity matrix between text and aligned vision tokens (termed Vision-Language Affinity Distillation) to enhance cross-modal alignment.

Comprehensive experiments on multiple vision understanding and parsing benchmarks indicate that EM-KD demonstrates performance improvements without altering the model architecture, validating the effectiveness. 
The EM-KD trained model outperforms previous Efficient MLLMs by a significant margin on both accuracy and efficiency, and when compared with previous MLLM distillation methods under broader evaluation settings, EM-KD consistently achieves superior performance.
We summarize our contributions as follows:
\begin{itemize}
    \item We propose EM-KD, a novel Knowledge Distillation framework designed to transfer visual comprehension and parsing capabilities from a vanilla teacher model to an Efficient MLLM.
    \item EM-KD overcomes the challenge of unbalanced vision tokens in the spatial dimension between the teacher and student with the Hungarian algorithm. For transferring more comprehensive knowledge, EM-KD introduces Vision-Language Affinity Distillation and Vision Semantic Distillation.
    \item Compared with previous Efficient MLLMs, the EM-KD trained model exhibits superior performance on both accuracy and efficiency without altering the model architecture.
\end{itemize}

\section{Related Work}
\label{sec:related_work}
\begin{figure*}[!th]
\centering
\includegraphics[width=0.96\textwidth]{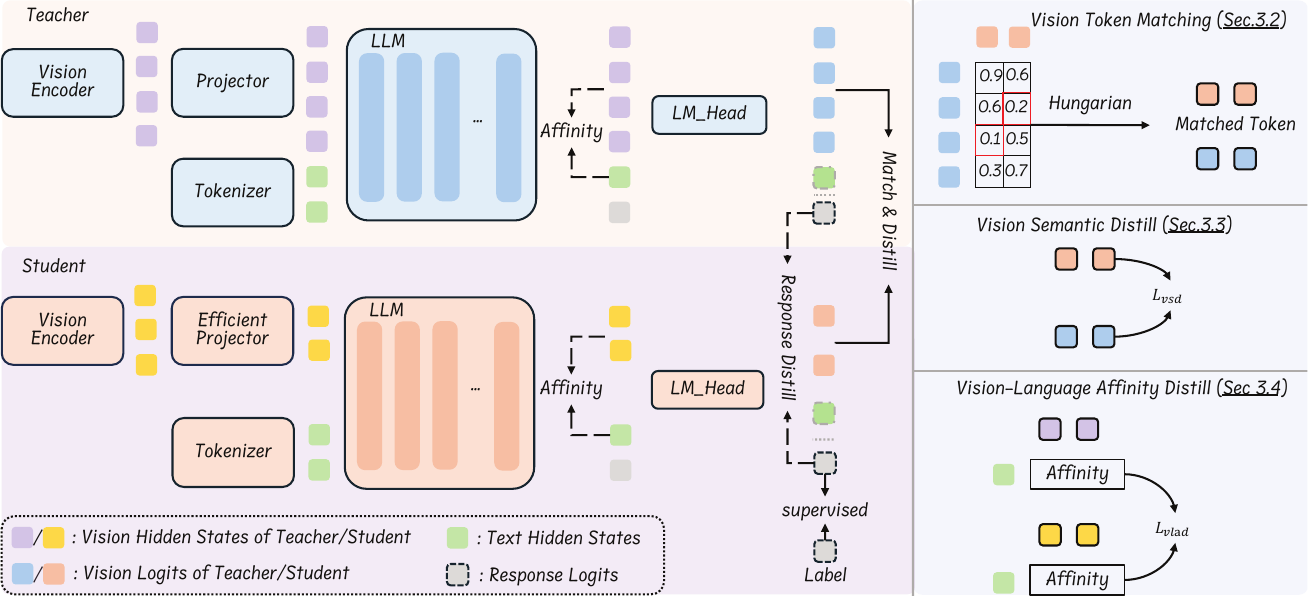} 
\caption{\textbf{Framework of the proposed EM-KD}. EM-KD consists of three key components: Vision Token Matching, Vision Semantic Distillation and Vision-Language Affinity Distillation. Vision Token Matching employs the Hungarian algorithm to resolve the unbalanced vision tokens problem between the teacher and student. Vision Semantic Distillation performs one-to-one knowledge transfer between matched tokens in logits space. Vision-Language Affinity Distillation further strengthens cross-modal alignment.}
\label{framework}
\end{figure*}
\subsection{Multimodal Large Language Models}
By integrating vision encoders into the LLM \cite{touvron2023llama, bai2023qwen, qwen2, liu2024deepseek, cai2024internlm2}, MLLMs gain the capability for visual understanding and parsing \cite{achiam2023gpt, alayrac2022flamingo, cambrian, li2023blip}.
LLaVA \cite{li2024llavaov, liu2024llavanext, liu2024visual, liu2024improved} series aligns vision features with text features through a vision projector, and feeds the combined representation into the LLM for processing.
Since the computational complexity of Transformer is quadratic to the length of the token, these methods face significant challenges in inference efficiency.

Efficient MLLMs reduce computational cost by compressing or directly pruning redundant vision tokens.
DeCo \cite{yao2024deco} employs adaptive average pooling for compact vision token compression, 
TokenPacker \cite{li2025tokenpacker} and Vision Remember \cite{feng2025vision} mitigate vision forgetting caused by token compression through vision feature resampling mechanisms.
Many methods focus on directly pruning redundant tokens without requiring post-training.
FastV \cite{chen2024image} analyzes attention sparsity to identify redundant vision tokens and proposes early-stage pruning in shallow decoder layers.
VisPruner \cite{zhang2025vispruner} strategically focuses on pruning vision encoders rather than LLM, preserving cross-modal alignment capabilities.

\subsection{Knowledge Distillation in MLLMs}
Many studies have introduced knowledge distillation techniques into the field of MLLMs. Based on the distillation targets, these approaches can be categorized into two types: LLM-style and MLLM-style.
LLM-style methods \cite{lee2025genrecal,ko2024distillm,distillm2,EPIC,minillm,shullavamod} mainly focus on response token distillation.
MiniLLM \cite{minillm} adopts the Reverse Kullback-Leibler Divergence (RKLD) to avoid the student model overestimating the teacher’s low-probability areas, making it more suitable for LLM distillation.
LLaVA-MOD \cite{shullavamod} incorporates distillation during the DPO phase, utilizing teacher guidance to mitigate hallucination in the smaller student model.
Although LLM-style methods offer greater flexibility, they fail to fully unleash MLLMs' comprehension capabilities due to their neglect of both the distinctive characteristics of vision features and vision-language correlations.

MLLM-style approaches \cite{xu2024llavadi,move-kd,cai2025llavakd,feng2025alignkd} primarily focus on distilling vision feature consistency and modeling vision-language correlations.
LLaVA-KD \cite{cai2025llavakd} aligns teacher and student visual representations by modeling the correlations between vision tokens.
Align-KD \cite{feng2025alignkd} focuses more on distillation between Efficient MLLMs, enhancing the expressive capability of smaller models through vision feature alignment.
However, both LLaVA-KD and Align-KD can only perform distillation between teacher and student models that share identical vision encoders and projectors.
Unlike these approaches, our proposed EM-KD targets a more general setting, where a fully pre-trained vanilla teacher model is used to distill efficient student models, accommodating scenarios with unbalanced vision tokens.

Due to the pages limitation, we provide more detailed comparison with related works in \textit{Supplementary Materials}.

\section{Method}
In this section, we first give a brief introduction to an Efficient MLLM, which serves as our baseline and student model.
And then, we introduce the proposed EM-KD framework, including three key components: Vision Token Matching, Vision-Language Affinity Distillation, and Vision Semantic Distillation. The framework of EM-KD is shown in Figure \ref{framework}.

\subsection{Efficient Baseline}
Representative MLLMs typically consist of three main components: (1) a vision encoder, (2) a vision projector, and (3) an LLM. 
The vision encoder, usually implemented as a Vision Transformer \cite{dosovitskiy2020vit} pretrained on large-scale datasets through vision-language contrastive learning approach, such as CLIP \cite{radford2021clip} and SigLip \cite{zhai2023sigmoid}, is responsible for extracting vision features from images.
The currently widely used vision projector is an MLP, which aligns vision features to the language feature space.
To reduce the number of vision tokens, we first compress vision features using Adaptive Average Pooling, followed by a two-layer MLP to form an efficient baseline, which also serves as our student model.
One of the primary causes of unbalanced vision tokens is the different vision encoders in the teacher and student. In particular, the student model employs an efficient encoder with token compression, resulting less vision tokens, while teacher retain more vision tokens from powerful encoder to keep the fine-grained understanding capability.
Finally, both vision tokens and text tokens are fed into a pre-trained LLM for processing and understanding. 
The MLLM then generates responses under the next-token prediction paradigm.

\subsection{Vision Token Matching}
As previously mentioned, the number of vision tokens differs between the teacher and student models in broader settings, and they are not one-to-one aligned in the spatial dimension.
The misalignment fundamentally disrupts traditional MLLM-style distillation methods, which typically rely on strict token-level correspondence between teacher and student.

Given the vision tokens $T_{v}^{t} \in \mathbb{R}^{N_{v}^{t}\times D}$, $T_{v}^{s} \in \mathbb{R}^{N_{v}^{s}\times D}$ belonging to teacher and student respectively, we must find a permutation $\delta \in \Theta_{N_{v}^{t}}$ with the lowest cost:
\begin{equation}
    \epsilon = \mathop{\arg\min}\limits_{\delta}\sum\mathcal{C}_{match}(T_{v}^{t}, T_{v}^{s}),
\end{equation}
where $N_{v}^{t}, N_{v}^{s}$ indicate the vision tokens length in teacher and student, $D$ is feature dimension, and $\mathcal{C}_{match}$ is the pair-wise matching cost with permutation $\delta$.
Inspired by DETR \cite{carion2020end}, in EM-KD, we treat this as a set bipartite matching task and employ Hungarian algorithm.
First, we decode the $T_{v}^{t}, T_{v}^{s}$ into vocabulary space via the LM head $\theta^{t}, \theta^{s}$ as the vision logits,
and then we compute the \textit{Manhattan distance} as the cost matrix $\mathcal{C}_{match} \in \mathbb{R}^{N_{v}^{t} \times N_{v}^{s}}$:
\begin{equation}
    \mathcal{C}_{match} = Manhattan(\theta^{t}(T_{v}^{t}),  \theta^{s}(T_{v}^{s})).
\end{equation}
Finally, the Hungarian algorithm running on the GPU is adopted to obtain the matching permutation $\delta$.
An alternative matching approach is to use 1D \textit{Adaptive Average Pooling} to reduce the longer vision tokens. 
However, this method loses visual information and fails to leverage the teacher model’s fine-grained understanding capability, leading to suboptimal accuracy (we will demonstrate in subsequent ablation experiments).

\subsection{Vision Semantic Distillation}
Considering the semantic richness (mentioned in the preliminary analysis in Sec.\ref{sec:intro}) of vision logits in the final layer, we propose using the \textit{reverse KL Divergence} tailored for text logits to transfer the vision knowledge.
Given the vision logits of the teacher and student models, and matching permutation $\delta$, we calculate the loss function as follows:
\begin{equation}
    \mathcal{L}_{\mathrm{vsd}}(\pi_\mathrm{S}; \pi_\mathrm{T}) = \mathbb{E}_{(x, y) \sim \pi_\mathrm{S}} \left[ \log \frac{\pi_\mathrm{S}(y \mid x)}{\pi_\mathrm{T}(y \mid x)} \right],
\end{equation}
where $\pi_\mathrm{S}(y \mid x)$ and $\pi_\mathrm{T}(y \mid x)$ mean the probability distribution of the matched vision logits condition on input $x$ for student and teacher.
Distilling logits instead of hidden states eliminates the need for an additional projection layer to align the dimensions of the teacher and student models, as they share the same vocabulary.
This is also a key benefit of Vision-Language Affinity Distillation.

\subsection{Vision-Language Affinity Distillation}
In traditional vision distillation methods, the primary focus is on modeling relation knowledge between vision features, and LLaVA-KD \cite{cai2025llavakd} also follows this principle.
We argue that in MLLMs, the relationship between vision and language should be prioritized over vision feature relationships, as it directly measures vision-language alignment and evaluates the semantic affinity.
Given the matched vision tokens $\hat{T}_{v}^{t}, \hat{T}_{v}^{s}\in \mathbb{R}^{N_v^{s} \times D}$ of the teacher and student models, we compute their \textit{cosine distance} with the corresponding language tokens $T_{l}^{t}, T_{l}^{s} \in \mathbb{R}^{N_t \times D}$ as affinity matrices $R_{t}, R_{s}$:
\begin{equation}
    R_{t} = Cos(\hat{T}_{v}^{t}, T_{l}^{t}),
    \
    R_{s} = Cos(\hat{T}_{v}^{s}, T_{l}^{s}),
    \
    \in \mathbb{R}^{N_v^{s} \times N_t},
\end{equation}
where $D$ is the hidden dimension and $N_t$ means the length of language tokens (in the SFT paradigm, they are equal in teacher and student).
Then we compute the \textit{Smooth L1} Loss between $R_{t}$ and $R_{s}$:
\begin{equation}
    \mathcal{L_\mathrm{vlad}} = smooth_{L1}(R_{t}, R_{s}).
\end{equation}
Although measuring affinity between logits (rather than hidden states) may seem like a more principled choice, the high dimensionality of logits would consume excessive GPU memory. 
We therefore adopt this memory-efficient alternative.

\subsection{Overall Training Loss}
\begin{algorithm}[!t]
\caption{Pseudocode of EM-KD in a PyTorch-like style}
\label{alg:emkd}
\begin{lstlisting}[language=Python, style=mystyle, mathescape]
# vhs_s, vhs_t: vision hidden states of student and teacher
# lhs_s, lhs_t: language hidden states of student and teacher
# rl_s, rl_t: response logits of student and teacher
# head_s, head_t: LM_head of student and teacher
    
L_disitll = as_tensor(0)
# decode the vision tokens as vision logits
vl_s, vl_t = head_s(vhs_s), head_t(vhs_t)
for data in batch:
  # no gradient when match vision logits
  with no_grad():
    # calculate Manhattan distance, same as Eq.(2)
    cost = cdist(vl_s, vl_t, p=1).detach()
    # Hungarian algorithm on GPU
    idx_s, idx_t = hungarian_gpu(cost).cpu()

  # distill matched vision logits, same as Eq.(3)
  L_vsd = reverse KLD(vl_s[idx_s], vl_t[idx_t])

  # affinity between vision and text hidden states same as Eq.(4)
  affinity_s = cosine_similarity(vhs_s[idx_s], lhs_s)
  affinity_t = cosine_similarity(vhs_t[idx_t], lhs_t)
    
  # distill loss between affinity, same as Eq.(5)
  L_vlad = smooth_l1_loss(affinity_s, affinity_t)
    
  L_disitll += (beta * L_vsd + gamma * L_vlad)

# distill loss and sft loss between response logits, same as Eq.(6)
L_rld = reverse KLD(rl_s, rl_t)
L_sup = CrossEntropy(rl_s, labels)

loss = alpha * L_sup + (1 - alpha) * L_rld + L_disitll/batch_size
\end{lstlisting}
\end{algorithm}
Following the common practice, in addition to the vision semantic loss $\mathcal{L}_{\mathrm{vsd}}$ and the vision-language affinity loss $\mathcal{L}_{\mathrm{vlad}}$, we further incorporate supervised loss $\mathcal{L}_{\mathrm{sup}}$ and distillation loss $\mathcal{L}_{\mathrm{rld}}$ applied to the auto-regressive response logits as follows:
\begin{equation}
\begin{split}
    & \mathcal{L}_{\mathrm{sup}}(\pi_\mathrm{S}) = \mathbb{E}_{(y_k \mid y_{<k}, x) \sim \pi_\mathrm{S}} \left[ \log \pi_\mathrm{S}(y_k \mid y_{<k}, x) \right], \\
    & \mathcal{L}_{\mathrm{rld}}(\pi_\mathrm{S}; \pi_\mathrm{T}) = \mathbb{E}_{(x, y_k) \sim \pi_\mathrm{S}} \left[ \log \frac{\pi_\mathrm{S}(y_k \mid y_{<k}, x)}{\pi_\mathrm{T}(y_k \mid y_{<k}, x)} \right],
\end{split}
\end{equation}
The overall loss can be formulated as:
\begin{equation}
    \mathcal{L} = \alpha\mathcal{L}_{\mathrm{sup}} + (1-\alpha)\mathcal{L}_{\mathrm{rld}} + \beta\mathcal{L}_{\mathrm{vsd}} + \gamma\mathcal{L_\mathrm{vlad}},
\end{equation}
where $\alpha, \beta$ and $\gamma$ are weights of each objective item to balance overall loss.
Algorithm \ref{alg:emkd} provides the pseudo code of EM-KD in PyTorch-like style.

\section{Experiments}
\subsection{Experimental Settings}
\subsubsection{Implementation Details.}
We choose the pretrained LLaVA-OneVision-SI \cite{li2024llavaov} in two different sizes (0.9B and 8B) as teacher models, and LLaVA-NeXT \cite{liu2024llavanext} equipped with the efficient projector as student models.
For the student model, we employ two scaled variants of Qwen2 \cite{qwen2} as the LLM paired with different-sized SigLip vision encoders \cite{zhai2023sigmoid}, constructing three baseline configurations with 0.6B and 8B parameters respectively.
Following AnyRes strategy proposed in \cite{li2024llavaov}, each image patch is processed by an efficient projector that compresses vision tokens to a maximum of 144 per patch.

We train the MLLM in two phases. 
\textit{Phase-1: Language-Image Alignment.} 
In this phase, we use the image-caption pairs in the CC-558K dataset \cite{liu2024visual} to train the efficient vision projector. 
\textit{Phase-2: Visual Instruct Tuning.} 
In this phase, the whole student model is trained.
The 779K mixture dataset \cite{liu2024llavanext} is used to enhance the MLLM's capability of vision understanding and instruction following.
We train all models in each phase for one epoch and the proposed EM-KD is only used in Phase-2.
The loss weights $\alpha, \beta$ and $\gamma$ are set to 0.5, 0.25 and 25 to balance the loss items.
The experiments are conducted on 8 $\times$ NVIDIA H20 GPUs.
Dur to the pages limitation, more model configuration and implementation details could be found in the \textit{Supplementary Materials}.
\begin{table*}[!ht]
\centering
\scriptsize
\setlength{\tabcolsep}{3.0pt}
\renewcommand{\arraystretch}{1.2}
\begin{tabular}{c|ccc|cccc|cccc|c|c}
\multicolumn{1}{c|}{\multirow{2}{*}{Methods}} & \multicolumn{3}{c|}{General} & \multicolumn{4}{c|}{Knowledge} & \multicolumn{4}{c|}{OCR\&Chart} & \multicolumn{1}{c|}{\multirow{2}{*}{\textbf{Average}$\uparrow$}} & \multicolumn{1}{c}{\multirow{2}{*}{\textbf{TTFT/ms}$\downarrow$}} \\ 
\multicolumn{1}{c|}{} & \multicolumn{1}{c}{GQA} & \multicolumn{1}{c}{$\text{MME}^\text{P}$} & \multicolumn{1}{c|}{RWQA} & \multicolumn{1}{c}{$\text{SQA}^\text{I}$} & \multicolumn{1}{c}{AI2D} & \multicolumn{1}{c}{$\text{MMMU}^\text{V}$} & \multicolumn{1}{c|}{MMStar} & \multicolumn{1}{c}{ChartQA} & \multicolumn{1}{c}{DocVQA} & \multicolumn{1}{c}{TextVQA} & \multicolumn{1}{c|}{OCRBench} & \multicolumn{1}{c|}{} & \multicolumn{1}{c}{} \\
\hline
\multicolumn{13}{c}{\textit{Vanilla Method without Vision Token Pruning or Compression}} \\
\hline
LLaVA-NeXT \cite{liu2024llavanext} & \underline{59.3} & 1203.7 & 48.2 & \underline{53.1} & \underline{52.3} & 30.2 & 34.3 & 57.6 & 61.9 & 47.3 & 38.6 & \textbf{49.4}\textcolor{gray}{($\Delta0.0$)} & \textbf{103.3}\textcolor{gray}{($\Delta0.0$)} \\
\hline
\multicolumn{13}{c}{\textit{Training-free Vision Token Pruning}} \\
\hline
FastV \cite{chen2024image}  & 56.3 & 1200.3 & 47.9 & 53.0 & 51.0 & \underline{31.0} & 33.2 & 51.5 & 57.4 & 43.5 & 31.6 & \textbf{47.0}\textcolor{teal}{(-2.4)} & \textbf{75.1}\textcolor{red}{(+28.2)} \\
\rowcolor{lightgray!20}
PyramidDrop \cite{xing2024pyramiddrop} & 57.3 & 1210.4 & 48.1 & \underline{53.1} & 51.3 & 29.8 & 34.4 & 52.3 & 58.4 & 45.0 & 33.9 & \textbf{47.7}\textcolor{teal}{(-1.7)} & \textbf{80.5}\textcolor{red}{(+22.8)} \\
VisPruner \cite{zhang2025vispruner} & 57.1 & 1205.6 & 48.2 & 52.6 & 51.2 & 30.7 & 34.5 & 52.6 & 58.2 & 44.8 & 34.7 & \textbf{47.7}\textcolor{teal}{(-1.7)} & \textbf{61.7}\textcolor{red}{(+41.6)} \\
\hline
\multicolumn{13}{c}{\textit{Training-base Vision Token Compression}} \\
\hline
\rowcolor{lightgray!20}
DeCo \cite{yao2024deco} & 57.7 & 1157.6 & 50.0 & 53.0 & 51.4 & 30.7 & 34.1 & 53.8 & 57.4 & 44.3 & 34.4 & \textbf{47.7}\textcolor{teal}{(-1.7)} & \underline{\textbf{54.9}}\textcolor{red}{(+48.4)} \\
TokenPacker \cite{li2025tokenpacker} & 57.4 & 1201.5 & 47.7 & 51.8 & 51.3 & 29.6 & 34.3 & 52.9 & 56.7 & 41.4 & 33.5 & \textbf{47.0}\textcolor{teal}{(-1.7)} & \textbf{61.0}\textcolor{red}{(+42.3)} \\
\rowcolor{blue!7}
\textbf{EM-KD (Ours)} & 57.8 & \underline{1226.5} & \underline{51.5} & \underline{53.1} & 51.5 & \underline{31.0} & \underline{36.1} & \underline{59.1} & \underline{64.1} & \underline{49.2} & \underline{39.7} & \underline{\textbf{50.4}}\textcolor{red}{(+1.0)} & \underline{\textbf{54.9}}\textcolor{red}{(+48.4)} \\
\end{tabular}
\caption{\textbf{Performance comparisons with various Efficient MLLMs.} When compared with other Efficient MLLMs, we all reduce the number of vision tokens to 144. We implement these methods under the same setting and test the TTFT on a single A100 GPU. The score of $\text{MME}^\text{P}$ is divided by 20 when calculating the average accuracy. The results of our method are highlighted with \colorbox{blue!7}{blue}. $\downarrow$ means that lower is better, $\uparrow$ means that higher is better. The best results are \underline{underlined}.
}
\label{tab:efficient_compare}
\end{table*}

\subsubsection{Evaluation.}
We conduct extensive experiments on 11 benchmarks to validate the understanding and parsing capabilities of the proposed method.
Specifically, we categorize all benchmarks into three distinct groups based on different focus areas:
(1) General Question Answer benchmarks include GQA \cite{gqa}, MME-Perception \cite{mme} and RealWorldQA \cite{rwqa}, 
(2) Comprehensive Knowledge Reasoning benchmarks include ScienceQA\_Image \cite{scienceqa}, AI2D \cite{ai2d}, MMMU \cite{mmmu} and MMStar \cite{mmstar},
(3) OCR\&Chart Parsing benchmarks include ChartQA \cite{chartqa}, DocVQA \cite{docvqa}, TextVQA \cite{textvqa} and OCRBench \cite{ocrbench}.
To compare the accuracy of MLLMs, we take the average scores on the whole benchmark.
For efficiency comparison, we evaluate the \textit{Time to First Token (TTFT)} latency on the RealWorldQA.

\subsection{Main Results}
\label{sec:main_results}
\begin{table*}[!ht]
\centering
\scriptsize
\setlength{\tabcolsep}{5.0pt}
\renewcommand{\arraystretch}{1.2}
\begin{tabular}{c|ccc|cccc|cccc|c}
\multicolumn{1}{c|}{\multirow{2}{*}{Methods}} & \multicolumn{3}{c|}{General} & \multicolumn{4}{c|}{Knowledge} & \multicolumn{4}{c|}{OCR\&Chart} & {\multirow{2}{*}{\textbf{Average}$\uparrow$}} \\ 
\multicolumn{1}{c|}{} & \multicolumn{1}{c}{GQA} & \multicolumn{1}{c}{$\text{MME}^\text{P}$} & \multicolumn{1}{c|}{RWQA} & \multicolumn{1}{c}{$\text{SQA}^\text{I}$} & \multicolumn{1}{c}{AI2D} & \multicolumn{1}{c}{$\text{MMMU}^\text{V}$} & \multicolumn{1}{c|}{MMStar} & \multicolumn{1}{c}{ChartQA} & \multicolumn{1}{c}{DocVQA} & \multicolumn{1}{c}{TextVQA} & \multicolumn{1}{c|}{OCRBench} & \\
\hline
\multicolumn{13}{c}{\textit{LLaVA-OneVision-SI-0.5B distill 0.6B Efficient Baseline}} \\
\hline
Efficient Baseline & 58.7 & 1157.6 & 50.0 & 53.4 & 51.8 & 30.7 & 33.8 & 53.8 & 56.2 & 44.3 & 34.3 & \textbf{47.7} \\
\rowcolor{lightgray!20}
MiniLLM \cite{minillm}  & 57.4 & 1247.3 & 51.5 & 53.9 & 51.4 & 29.0 & 35.0 & 57.3 & 62.4 & 47.4 & 36.8 & \textbf{49.5} \\
LLaVA-KD$^{\ast}$ \cite{cai2025llavakd} & 57.5 & 1203.8 & 52.0 & 52.6 & 52.6 & 29.9 & 34.5 & 57.3 & 62.7 & 48.2 & 39.0 & \textbf{49.7} \\
\rowcolor{blue!7}
\textbf{EM-KD (Ours)} & 57.8 & 1226.5 & 51.5 & 53.0 & 51.5 & 31.0 & 36.1 & 59.1 & 64.1 & 49.2 & 39.7 & \textbf{50.4}\\
\hline
\multicolumn{13}{c}{\textit{LLaVA-OneVision-SI-7B distill 8B Efficient Baseline}} \\
\hline
Efficient Baseline & 61.4 & 1455.0 & 61.2 & 70.3 & 72.1 & 33.8 & 43.3 & 69.7 & 76.9 & 65.4 & 51.6 & \textbf{61.7} \\ 
\rowcolor{lightgray!20}
MiniLLM \cite{minillm} & 62.6 & 1478.4 & 61.8 & 51.0 & 71.7 & 71.7 & 38.3 & 45.1 & 70.4 & 77.0 & 64.2 & \textbf{62.5} \\
LLaVA-KD$^{\ast}$ \cite{cai2025llavakd} & 62.7 & 1465.3 & 61.8 & 72.1 & 71.5 & 38.5 & 45.3 & 69.5 & 75.1 & 63.9 & 51.3 & \textbf{62.3} \\
\rowcolor{blue!7}
\textbf{EM-KD (Ours)} & 65.1 & 1514.9 & 62.2 & 77.4 & 72.4 & 38.9 & 46.0 & 69.6 & 74.3 & 64.1 & 51.7 & \textbf{63.4} \\
\end{tabular}
\caption{\textbf{Performance comparisons with various MLLM distillation approaches.} Efficient Baseline indicates the model was only trained with SFT supervision. $^{\ast}$ means we integrate the Vision Token Matching into LLaVA-KD. The score of $\text{MME}^\text{P}$ is divided by 20 when calculating the average accuracy. We implement these methods under the same setting. The results of our method are highlighted with \colorbox{blue!7}{blue}. $\uparrow$ means that higher is better.}
\label{tab:distill_compare}
\end{table*}
\subsubsection{Comparison with other Efficient MLLMs.}
The ultimate goal of our work is to train an Efficient MLLM. To this end, we conduct comparisons with representative approaches for training-free vision token pruning (including FastV \cite{chen2024image}, PyramidDrop \cite{xing2024pyramiddrop}, and VisPruner \cite{zhang2025vispruner}) and training-based vision token compression (including DeCo \cite{yao2024deco}, TokenPacker \cite{li2025tokenpacker}).
All methods are trained on the same datasets, and employ Qwen2-0.5B as the LLM and SigLip-base as the vision encoder.

For training-free vision token pruning approaches, we first train the LLaVA-NeXT \cite{liu2024llavanext} model without pruning, serving as their baseline. And then employ the FastV \cite{chen2024image}, PyramidDrop \cite{xing2024pyramiddrop}, and VisPruner \cite{zhang2025vispruner} to prune the vision tokens.
For DeCo \cite{yao2024deco} and TokenPacker \cite{li2025tokenpacker}, we directly replace the vision projector and train the model.

As shown in Table \ref{tab:efficient_compare}, the EM-KD trained model achieves 50.4 average score, outperforming all previous Efficient MLLMs by a significant margin, including FastV, PyramidDrop, TokenPacker and DeCo. 
An interesting finding is that our method still surpasses LLaVA-NeXT (49.4) with +1.0 accuracy improvement. Notably, LLaVA-NeXT is not an Efficient MLLM—it retains up to 576 tokens per image patch, resulting in higher latency, whereas EM-KD uses only up to 144 tokens per patch.
This demonstrates that EM-KD produces models with fewer vision tokens (\textit{i.e.}, more efficient inference) while achieving superior performance—fully validating the effectiveness of our methods.

Furthermore, we conduct efficiency comparisons between all methods.
Compared with LLaVA-NeXT \citep{liu2024llavanext} (103.3ms TTFT), which does not compress the vision tokens, all other methods achieve efficiency improvements due to the short embedding sequence.
The acceleration effect of training-free vision token pruning methods is generally inferior to that of training-based vision token compression approaches.
This discrepancy is because training-free methods rely on attention maps to select important tokens, which is fundamentally incompatible with mature acceleration operators such as Flash Attention and Scaled Dot-Product Attention.
Among all methods, the model trained with EM-KD achieves the fastest speed (54.9ms), benefiting from the most compact network architecture.

\subsubsection{Comparison with other Knowledge Distillation methods.}
To demonstrate the effectiveness of EM-KD, we make fair comparisons with two previous  representative methods: MiniLLM \cite{minillm} and LLaVA-KD \cite{cai2025llavakd}. 
It is worth noting that LLaVA-KD cannot directly distill efficient student models due to the unbalanced vision tokens. 
We address this by integrating the same Vision Token Matching, which is also a key contribution in this work, in EM-KD to it.

As shown in Table \ref{tab:distill_compare}, when using LLaVA-OneVision-SI-0.5B (actually 0.9B) to distill 0.6B efficient baseline, all distillation methods outperform the SFT-only baseline, validating our core hypothesis: MLLM distillation can not only perform well under varying model parameter scales, but also work across different vision token quantities.
What's more, EM-KD achieves the highest average score, surpassing the baselines and other distillation methods.
Specifically, EM-KD demonstrates consistent improvements across most tasks, such as DocVQA (64.1 \textit{v.s.} 56.2 baseline), ChartQA (59.1 \textit{v.s.} 53.8 baseline), and MMStar (36.1 \textit{v.s.} 33.8 baseline). 
Compared to the LLM-style representative method MiniLLM \cite{minillm}, EM-KD surpass it by 0.9 (50.4 \textit{v.s.} 49.5).
This occurs because MiniLLM solely optimizes response tokens while disregarding the rich visual semantics information in vision tokens.
Compared to the stronger competitor LLaVA-KD \cite{cai2025llavakd}, because considering vision-language affinity, EM-KD yields a 0.7 point improvement in average score (50.4 \textit{v.s.} 49.7). 
When we scale up the teacher and the student models, using LLaVA-OneVision-SI-7B to distill 8B efficient baselines, EM-KD exhibits consistent performance gains, outperforming two powerful counterparts.

\subsection{Ablation Study}
We conduct a number of ablation experiments to verify the effectiveness of each component in our EM-KD framework.
Unless specified, we report the results on the 0.6B efficient student with the LLaVA-OneVision-SI-0.5B serving as teacher, and the experiment settings are kept as the same as the Sec \ref{sec:main_results}.
\begin{table}[!tb]
    \centering
    \begin{subtable}[t]{0.46\textwidth} 
        \centering
        \scriptsize
        \setlength{\tabcolsep}{1.0pt}
        \renewcommand{\arraystretch}{1.2}
        \begin{tabular}{c|ccc|ccc|c}
            & VLAD & VSD & RLD & {General} & {Knowledge} & {OCR\&Chart} & \textbf{Average}$\uparrow$ \\ 
            \hline
            Baseline & & & & 55.5 & 42.4 & 47.2 & \textbf{47.7}\textcolor{gray}{($\Delta0.0$)} \\ 
            \hline
            \rowcolor{lightgray!20}
            \cellcolor{white}\multirow{3}{*}{EM-KD} & \cmark & & & 56.1 & 41.1 & 49.6 &\textbf{48.4} \textcolor{red}{({+0.7})} \\
            & \cmark & \cmark & & 56.6 & 42.1 & 51.5 &\textbf{49.5} \textcolor{red}{({+1.8})} \\
            \rowcolor{blue!7}
            \cellcolor{white} & \cmark & \cmark & \cmark & 56.9 & 42.9 & 53.0 & \textbf{50.4} \textcolor{red}{({+2.7})} \\
        \end{tabular}
        \caption{Ablation study of \textbf{key components}.}
        \label{tab:key}
    \end{subtable} 
    \begin{subtable}[t]{0.45\textwidth} 
        \centering
        \scriptsize
        \setlength{\tabcolsep}{1.0pt}
        \renewcommand{\arraystretch}{1.2}
        \begin{tabular}{c|ccc|c}
            {Matching Methods} & {General} & {Knowledge} & {OCR\&Chart} & {\textbf{Average}$\uparrow$} \\ 
            \hline
            \rowcolor{lightgray!20}
            Average Pooling             & 55.8 & 41.5 & 50.2 & \textbf{48.6} \textcolor{teal}{(-1.8)} \\
            Hungarian by Hidden States  & 56.6 & 41.8 & 51.5 & \textbf{49.4} \textcolor{teal}{(-1.0)} \\
            \rowcolor{blue!7}
            Hungarian by Logits & 56.9 & 42.9 & 53.0 & \textbf{50.4}\textcolor{gray}{($\Delta0.0$)} \\
        \end{tabular}
        \caption{Experimental results with various \textbf{matching methods} in EM-KD.}
        \label{tab:match}
    \end{subtable}
    \begin{subtable}[t]{0.45\textwidth} 
        \centering
        \scriptsize
        \setlength{\tabcolsep}{3.0pt}
        \renewcommand{\arraystretch}{1.2}
        \begin{tabular}{c|ccc|c}
           {Distill Objects of VSD} & {General} & {Knowledge} & {OCR\&Chart} & {\textbf{Average}$\uparrow$} \\ 
           \hline
           {Hidden States} & 56.4 & 42.1 & 51.6 & \textbf{49.5} \textcolor{teal}{(-0.9)} \\
           \rowcolor{blue!7}
           {Logits} & 56.9 & 42.9 & 53.0 & \textbf{50.4}\textcolor{gray}{($\Delta0.0$)} \\
        \end{tabular}
        \caption{Experimental results with different \textbf{distillation objects of VSD} in EM-KD.}
        \label{tab:object}
    \end{subtable}
    \label{tab:ablation}
\caption{Ablation studies. The default setting is marked in \colorbox{blue!7}{blue}. $\uparrow$ means that higher is better.}
\end{table}
\subsubsection{Key Components.}
As shown in Table \ref{tab:key}. We evaluate the impact of three key components: VLAD, VSD, and RLD, on the performance across different task categories, including General, Knowledge and OCR\&Chart, as well as the overall average score.
Starting from the baseline, which achieves an average score of 47.7, we observe that introducing the VLAD alone leads to an improvement, raising the average score to 48.4 (+0.7). 
Adding the VSD on top of VLAD further improves the average score to 49.5 (+1.8 compared to baseline)
When all three components (VLAD, VSD, and RLD) are combined, the model achieves the best performance with an average score of 50.4 (+2.7 compared to baseline). Significant improvements are observed both in General benchmarks (from 55.5 to 56.9) and OCR\&Chart (from 47.2 to 53.0) benchmarks.

These results confirm that each component in our proposed EM-KD brings complementary benefits, and their joint application leads to the most substantial performance gains. The ablation study clearly demonstrates the effectiveness and necessity of all key components to achieve the best results.
\subsubsection{Matching Methods.}
Table \ref{tab:match} presents the experimental results of different matching methods in the EM-KD, including Average Pooling, Hungarian matching by Vision Hidden States, and Hungarian matching by Vision Logits.
Average Pooling achieves the lowest average score of 48.6.
Hungarian matching by Vision Hidden States slightly improves the performance, reaching an average score of 49.4.
When using Vision Logits to match the teacher and student, EM-KD achieves the best average score of 50.4.
This experiment demonstrates that the Hungarian algorithm enables more precise matching, significantly enhancing the performance of distillation methods.
Besides, matching based on Vision Logits is more beneficial because they possess explicit semantics, and distance metrics measured in vocabulary space give more accurate results than those in low-dimensional Hidden States.
\subsubsection{Distillation Objects of VSD.}
Table \ref{tab:object} compares the effect of different distillation objects for VSD: (1) distilling Vision Hidden States and (2) distilling Vision Logits.
When distilling Hidden States, the model achieves an average score of 49.5.
Distilling Logits further improves the performance, yielding an average score of 50.4.
This result suggests that using Logits as the distillation target is more effective than using Hidden States, consistent with the findings in the previous experiment.
The vision logits encapsulate the distilled knowledge and lead to better overall generalization on all benchmarks, especially in the OCR\&Chart dataset. 

\section{Conclusion}
In this paper, we propose EM-KD, a novel paradigm that employs a well-pretrained teacher to distill Efficient MLLM.
EM-KD addresses the visual token imbalance issue via the Hungarian algorithm, establishing strict token-level correspondence.
Building upon the optimal assignment, we further introduce Vision-Semantic Distillation and Vision-Language Affinity Distillation to transfer more knowledge from the teacher to the student.
We hope this work will draw the community’s attention to Efficient MLLMs.

\section{Acknowledgments}
This work is supported by Shenzhen Science and Technology Program under project [JCYJ20230807114659029]. 
This work is supported by the National Natural Science Foundation of China under No. 62276061 and 62436002.
This work is also supported by Research Fund for Advanced Ocean Institute of Southeast University (Major Program MP202404).
This research work is supported by the Big Data Computing Center of Southeast University.

\bibliography{aaai2026}

\end{document}